\documentclass[11pt,a4paper,table]{article} 
\pdfoutput=1
\usepackage[hyperref]{acl2020}
\usepackage{times}
\usepackage{latexsym}

\usepackage{microtype}
\usepackage{graphicx}
\usepackage{booktabs}
\usepackage{array}
\usepackage{arydshln}
\usepackage{verbatim}

\aclfinalcopy 
\def\aclpaperid{196} 

\title{Small Batch Sizes Improve Training of Low-Resource Neural MT}

\author{\`{A}lex R. Atrio$^{1,2}$ \and Andrei Popescu-Belis$^{1,2}$\\[8pt] 

\begin{tabular}{cc}
$^1$HEIG-VD / HES-SO  & \ $^2$EPFL \\\
Yverdon-les-Bains & Lausanne \\
Switzerland & ~~~~Switzerland~~~~ \\
\end{tabular}\\
\{alejandro.ramirezatrio, andrei.popescu-belis\}@heig-vd.ch
}

\date{}

\begin{document}
\maketitle
\begin{abstract}
We study the role of an essential hyper-parameter that governs the training of Transformers for neural machine translation in a low-resource setting: the batch size. Using theoretical insights and experimental evidence, we argue against the widespread belief that batch size should be set as large as allowed by the memory of the GPUs.  We show that in a low-resource setting, a smaller batch size leads to higher scores in a shorter training time, and argue that this is due to better regularization of the gradients during training.
\end{abstract}


\section{Introduction}

Training Transformers for low-resource neural machine translation (NMT), i.e.\ when only small parallel corpora are available, raises the challenge of finding optimal hyper-parameters. While several fixed configurations of the Transformer \citep{vaswani2017_attention} have been empirically validated by the community, such as `Base' or `Big', the settings of many other hyper-parameters rely on tips from practitioners. However, these values are not always suitable to low-resource settings, and systematic studies in these settings are rare \citep{araabi-monz-2020-optimizing,van2020optimal}.

In this paper, we show that the best values of a hyper-parameter that is essential for training, namely \emph{batch size}, differ in low-resource settings from those commonly accepted when larger data sets are available.  We analyze the role of small batch sizes, inspired by studies in computer vision \citep{keskar2016}, and then pinpoint empirically the optimal trade-off between a high batch size (for efficiency) and a small one (for regularization).  Although large batch sizes were found to lead to higher-quality models in experiments with high-resource NMT \citep{Popel_2018_tips,xu-etal-2020-dynamically}, we show here that smaller batch sizes can outperform the latter, likely due to a regularizing effect in the gradient update.  Moreover, we show that this finding is invariant to changes in tokenization methods.

The paper is organized as follows.  In Section~\ref{sec:SOTA-ML}, we discuss batch size from a machine learning perspective, showing why smaller values of batch size may act as regularizers.  Then, in Section~\ref{sec:SOTA-NMT}, we review studies of hyper-parameters in NMT.  In Section~\ref{sec:settings}, we present the parameters of our Transformer and the data from the WMT 2020 Low-resource task \citep{fraser-2020-findings} and other sources 
that we use in our experiments. In Section~\ref{sec:optimal-batch-size}, we provide empirical evidence that smaller batch sizes are preferable in low-resource settings.


\section{ML Perspective on Batch Size}
\label{sec:SOTA-ML}

Machine learning theory argues that performing back-propagation with large batch sizes leads to better optimization, because the estimates of the gradients are more accurate. Conversely, using small batches during training leads to noisier gradient estimations, i.e.\ with a larger variance in comparison to the gradient computed over the entire training set. 
Still, one advantage of small batch sizes is that they are more likely to make  parameters converge towards flatter minima of the loss \citep[Chapter~8.1.3]{Goodfellow-et-al-2016}, as explained below.  Such flatter minima have better generalization capacities, i.e.\ they maintain performance when presented with a new test set.

\citet{keskar2016} define a \emph{flat minimizer} -- as opposed to a \emph{sharp} one -- as a point in the parameter space that is a local minimum of the loss function, and where this function varies slowly in a relatively large neighborhood. \citet{keskar2016} point to the following \emph{generalization gap}: training with large batch sizes tends to converge towards sharp minimizers, which offer poorer generalization capacities. Conversely, \emph{small batch sizes allow convergence towards flat minimizers}, which are likely to generalize better. Thus, smaller batch sizes have \emph{exploration abilities}: the search is more likely to exit the basins of sharp minimizers, and to tend towards flat minimizers, from where noise will not cause them to exit.

Since a sharp minimizer requires high precision to be described, unlike a flat one, the more noise there is in the gradient, the more unlikely it is that the parameters will converge towards a sharp minimizer.  
This is precisely the contribution of a smaller batch size: introduce noise in the gradient estimation. According to this theoretical view, above a certain threshold of the batch size, the generalization capacities of a model deteriorate.  The threshold depends on several hyper-parameters, including the batch size.  Its role has not been fully settled yet, with observations and conclusions varying widely across studies \citep{dinh2017sharp, hoffer_trainlonger2017, goyal2018accurate, li2018visualizing, kawaguchi2020generalization}. Moreover, these studies are on image data sets, with fully connected or with convolutional NNs, which differ substantially from NMT settings.


\section{The Role of Batch Size in Neural MT}
\label{sec:SOTA-NMT}

Several recent studies in NMT have considered batch size among other hyper-parameters, but they have either been in high-resource settings \citep{Popel_2018_tips,xu-etal-2020-dynamically} or have given only marginal attention to batch size \citep{sennrich-zhang-2019-revisiting,araabi-monz-2020-optimizing}.

\citet{Popel_2018_tips} reported that BLEU scores increased with batch size (including when using more GPUs) in a Transformer-based NMT system, although with diminishing returns, recommending in particular that ``batch size should be set as high as possible''. Their experiments were performed using mainly two datasets, with respectively 58M and 15M sentence pairs. It thus remains an open question whether their findings regarding batch size also apply when much less training data is available.

\citet{sennrich-zhang-2019-revisiting} experimented with a recurrent network in a low-resource setting and found that smaller batch sizes were beneficial, along with other forms of regularization. They experimented with two batch sizes of 4,000 and 1,000 tokens, and observed improvements with the latter of 0.30 and 0.04 BLEU points on data sets with 5k and 160k sentence pairs, respectively.  It is difficult to predict from these results what the optimal batch size is for Transformer-based NMT.

\citet{araabi-monz-2020-optimizing} studied the role of 15 hyper-parameters of the Transformer, with several sizes of low-resource datasets.  For the largest training sizes tested (80k and 165k sentence pairs), larger batch sizes improved performance, with respectively 8,192 and 12,288 versus 4,096 for the other sizes.  For lower training sizes, smaller batch sizes did not improve performance, which the authors explain by Transformer's need for larger batches.  In our view, an alternative explanation is the order of optimization of the hyper-parameters (a grid search in which they optimize one hyper-parameter at a time):  batch size is \#12 out of 15, so by the time several sizes are compared, regularization has already been introduced in the model by dropouts on words, activation, and layers.  Late optimization of batch size, of warmup steps (\#14) or of learning rate (\#15) cannot properly determine their regularizing effects.

\citet{xu-etal-2020-dynamically} proposed to compute gradients while accumulating minibatches, and observed that increasing batch size stabilizes gradient direction up to a certain point, after which it starts to fluctuate.  They used this criterion to dynamically adjust batch sizes while training.  In their experiments with large training sets (4.5M and 36M sentence pairs), their average batch size was around 26k on two GPUs, and never lower than 7k.  Their observations on the gradient direction as more minibatches are accumulated are consistent with the findings of \citet{Popel_2018_tips} who see diminishing returns when increasing batch size.

\section{Datasets and Systems}
\label{sec:settings}

We train NMT systems with two low-resource parallel corpora, listed in the first two lines of Table~\ref{tab:data}: the Upper Sorbian (HSB) to German (DE) training data of the WMT 2020 Low-Resource Translation Task \citep{fraser-2020-findings} and a low-size excerpt of the German to English News Commentary v13 \citep{bojar-etal-2018-findings}, from which we randomly sampled 60k parallel lines. For the HSB-DE models, we also use the development and test sets provided by the WMT 
2020 and 2021 Low-Resource Translation Task \citep{fraser-2021-findings}, each consisting of 2k sentences, and for DE-EN we sample a development set and a test set from the original corpus, with 2k sentences each as well. We apply a common filtering process for all data used: we delete from all our data the sentences that are not between 2 and 300 words long, with resulting numbers of lines shown in Table~\ref{tab:data}.

\begin{table}[t]
\centering
\footnotesize
\begin{tabular}{llrrr}
\hline
\textbf{Dataset} & \textbf{Lang.} & \textbf{Orig.} & \textbf{Filt.} & \textbf{$\Delta$\%}   \\
\hline \hline
WMT20 Low-res.\     & HSB-DE & 60k  & 59.8k & 0.29  \\
News Comm.\ v13     & DE-EN  & 60k  & 59.9k & 0.20 \\
\hdashline
Sorbian Institute   & HSB & 339k & 339k & 0.00  \\
Witaj               & HSB & 222k & 220k & 0.84 \\
Web                 & HSB & 134k & 121k & 9.98 \\
Europarl v8         & DE  & 2.2M & 2.2M & 0.79 \\
News Comm.\ v15     & DE  & 422k & 411k & 2.58 \\
JW300               & DE  & 2.3M & 2.2M & 4.44 \\
Europarl v3         & DE  & 790k & 785k & 0.69 \\
Europarl v3         & EN  & 790k & 782k & 1.07 \\
\hline
\end{tabular}
\caption{Numbers of lines in the original and filtered corpora used in our experiments.  HSB stands for Upper Sorbian and $\Delta$\% for the proportion of lines filtered out.  The only \textit{parallel} corpora used for training NMT are the first two ones; the other corpora are only used to train the SentencePiece model.} 
\label{tab:data}
\end{table}


We build subword vocabularies using the Unigram LM model \citep{P16-1162, Kudo2018-xx} as implemented in SentencePiece\footnote{\url{https://github.com/google/sentencepiece}}, with the monolingual corpora from Table~\ref{tab:data}.  We train a shared model for HSB-DE with a vocabulary of 32k pieces, character coverage of 0.98, \textit{nbest}\,=\,1 and \textit{alpha}\,=\,0.  The HSB data adds up to 740k sentences, and we sample 680k sentences from three DE corpora, and add them to the 60k sentences from the DE side of the parallel HSB-DE corpus.  To train the SentencePiece model for the DE-EN, for comparison purposes, we treat German as a low-resource language, and sample 680k lines of English and German from Europarl v3 \citep{TIEDEMANN12.463}, which we combine respectively with the 60k lines extracted from the DE-EN parallel corpus.

We use the Transformer-Base \citep{vaswani2017_attention} in the implementation provided by OpenNMT \citep{P17-4012,klein-etal-2020-opennmt}, with the parameters given in Appendix~A.  Unless otherwise specified, we follow OpenNMT-py's recommended values for the hyper-parameters.\footnote{\url{https://opennmt.net/OpenNMT-py/examples/Translation.html}}

When using several GPUs with gradient accumulation, each GPU processes several batches, which are then accumulated across all GPUs and used to update the model at each step. Therefore, the \textit{effective batch size} is $B \times G \times A$, where $B$ is the individual batch size, $G$ is the number of GPUs and $A$ the number of accumulated batches, and differs from the \texttt{batch\_size} hyper-parameter $B$. We train all models on two GeForce RTX 1080Ti GPUs with 11 GB of memory each and accumulate gradients over two minibatches ($A = 2$), following OpenNMT-py's recommendation. Therefore, the \texttt{batch\_size} parameter is  not our effective batch size, which is four times larger. Throughout this work, we will refer to batch size $B$ as the \texttt{batch\_size} parameter, and report true \emph{epochs}, which we define as computed with the effective batch size as $S \times B_\mathit{eff} / N$, for $S$ training steps, $B_\mathit{eff}$ effective batch size, and $N$ number of source tokens in the training set.

Following OpenNMT-py's recommendations, we set the Adam hyper-parameters at $ \beta_1 = 0.9,\ \beta_2 = 0.998,\ \epsilon = 10^{-8} $ and apply at each step a scaling factor of two to Noam's learning rate schedule, setting warmup steps to 8k.  Translations are generated with a beam width of seven, with an ensemble of the last four saved checkpoints. We report BLEU scores \citep{papineni-etal-2002-bleu} obtained with SacreBLEU \citep{post-2018-call} on detokenized text.

\section{Experimental Results}
\label{sec:optimal-batch-size}

\begin{figure}
\centering 
\includegraphics[width=0.99\columnwidth]{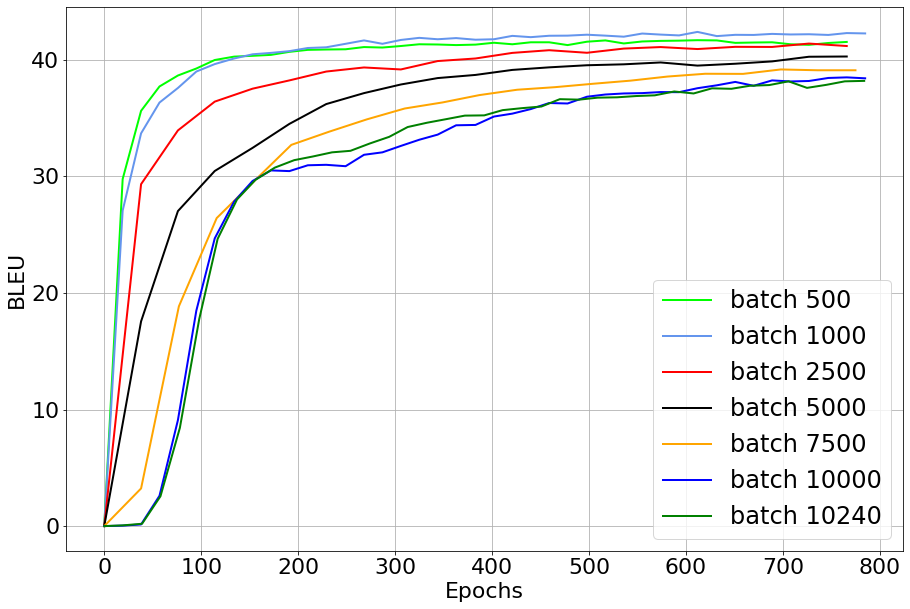}
\caption{BLEU scores on the test set for HSB-DE models trained with different batch sizes.}
\label{fig:hsb-de_bs_comparison} 
\end{figure}

To study the impact of batch sizes in a low-resource setting, we train various HSB-DE and DE-EN models for 700 epochs with the following batch sizes: 100, 250, 500, 1,000, 2,500, 5,000, 7,500, 10,000, and 10,240 (this is the largest one that fits in our GPU memory).

\subsection{NMT Performance}

\newcolumntype{A}{p{0.0275\textwidth}}
\newcolumntype{B}{p{0.035\textwidth}}

\begin{table*}[t]
\footnotesize
\centering
\begin{tabular}{r|AAAB|AAAB|AAAB|AAAB}
\toprule
    \textbf{Batch} & \multicolumn{8}{c|}{\textbf{HSB-DE}}  &\multicolumn{8}{c}{\textbf{DE-EN}}\\
    \textbf{Size} & \multicolumn{4}{c|}{dev} & \multicolumn{4}{c|}{test} & \multicolumn{4}{c|}{dev} & \multicolumn{4}{c}{test}\\
    & Xent & B{\scriptsize LEU} & chrF & TER & Xent & B{\scriptsize LEU} & chrF & TER & Xent & B{\scriptsize LEU} & chrF & TER & Xent & B{\scriptsize LEU} & chrF & TER\\
    \hline
    500 & 0.03 & \cellcolor{gray!20}{48.12} & 71.13 & 37.35 & 0.03 & \cellcolor{gray!20}{41.53} & 67.34 & 43.84 & 0.11 & \cellcolor{gray!20}{37.35} & 58.04 & 54.54 & 0.11 & 37.72 & 58.35 & 54.60\\
    
    1,000 & \textbf{0.02} & \textbf{49.23} & \textbf{72.07} & \textbf{36.35} & \textbf{0.02} & \textbf{42.26} & \textbf{67.93} & \textbf{43.16} & 0.05 & \cellcolor{gray!20}{\textbf{38.03}} & \textbf{59.39} & \textbf{52.91} & 0.05 & \textbf{38.67} & \textbf{59.68} & \textbf{52.71}\\
    
    2,500 & 0.03 & \cellcolor{gray!20}{48.28} & 71.63 & 37.02 & 0.03 & \cellcolor{gray!20}{41.18} & 67.36 & 44.02 & \textbf{0.04} & 33.83 & 56.70 & 56.27 & \textbf{0.04} & 35.51 & 57.76 & 55.47\\
    
    5,000 & 0.03 & 46.99 & 70.74 & 38.05 & 0.03 & 40.28 & 66.62 & 45.24 & 0.05 & \cellcolor{blue!10}{32.47} & 55.20 & 57.88 & 0.05 & \cellcolor{blue!10}{33.97} & 56.16 & 57.08\\
    
    7,500 & 0.03 & \cellcolor{blue!10}{46.05} & 70.29 & 38.87 & 0.03 & 39.10 & 65.94 & 46.18 & 0.05 & \cellcolor{blue!10}{32.67} & 55.99 & 57.67 & 0.05 & \cellcolor{blue!10}{33.80} & 56.72 & 57.21\\
    
    10,000 & 0.04 & 44.61 & 69.19 & 40.00 & 0.04 & \cellcolor{blue!10}{38.41} & 65.67 & 46.45 & 0.05 & \cellcolor{lime!10}{31.84} & 55.20 & 58.35 & 0.05 & \cellcolor{blue!10}{33.50} & 56.14 & 57.63\\
    
    10,240 & 0.04 & \cellcolor{blue!10}{45.59} & 70.12 & 39.26 & 0.04 & \cellcolor{blue!10}{38.19} & 65.39 & 46.79 & 0.06 & \cellcolor{lime!10}{31.49} & 55.00 & 58.65 & 0.06 & 33.03 & 55.78 & 58.07\\
    
\bottomrule
\end{tabular}
\caption{Loss and scores for models trained for 700 epochs with various batch sizes for HSB-DE and DE-EN directions.  All differences in BLEU on the dev and test sets are statistically significant at the 95\% level, except for the pairs in similar colors.
}
\label{tab:bs_comparison_60k}
\end{table*}

NMT performance on the HSB-DE test set throughout the training is shown in Figure~\ref{fig:hsb-de_bs_comparison}, with BLEU scores depending on the number of epochs. The evolution depending on real training time (wall time) is similar in terms of rankings.  Thus, the following analysis holds whether we train the models for the same amount of epochs or of hours.

The final scores on the development and test sets are given in Table~\ref{tab:bs_comparison_60k}, sorted by batch sizes.  We provide first the actual loss of the model (`Xent' for cross-entropy), and then three typical NMT scores: BLEU \citep{papineni-etal-2002-bleu}, chrF \citep{popovic-2015-chrf} and Translation Error Rate \citep{snover-2006-ter}.  The 100 and 250 batch size models did not reach BLEU scores significantly above zero, and are not included among the results in the table.

We test the statistical significance of the differences between each score and the others, with 95\% confidence, using the paired bootstrap resampling tool from SacreBLEU \citep{post-2018-call}.\footnote{\url{github.com/mjpost/sacrebleu}
with the signature nrefs:1$|$bs:1000$|$seed:12345$|$case:mixed$|$eff:no $|$tok:13a$|$smooth:exp$|$version:2.0.0.}
All differences between higher and lower BLEU scores are statistically significant,
except the pairs highlighted in similar colors in Table~\ref{tab:bs_comparison_60k}.\footnote{The difference in BLEU between the following pairs is not significant.
For HSB-DE, 2,500 vs.\ 500, and 10,240 vs.\ 10,000, on the test set; and 2,500 vs.\ 500, and 7,500 vs.\ 10,240 on the dev set.
For DE-EN these are 7,500 vs.\ 5,000, and 10,000 vs.\ 7,500, on the test set; and 1,000 vs.\ 500, 5,000 vs.\ 7,500, and 10,000 vs 10,240, on the dev set.}
The best NMT scores, which are always obtained with a batch size of 1,000, are significantly higher than all the other ones, including those obtained with the largest possible batch sizes for our GPU (10,000 or 10,240). We thus select two values for further experiments: a batch size of 1,000 as our highest-scoring model, and one of 10,000 as the maximum allowed by our GPU memory.  A simple ratio of 10 holds between the two values.

These empirical results are contrary to those from \citet{Popel_2018_tips}, who observe that increasing the batch size for Transformer-Base produces higher scores, although with diminishing returns after a certain threshold. We hypothesize that the main explanation is the difference between the amounts of training data: in our low-resource setting, we use 60k sentences, while \citet{Popel_2018_tips} use 57M sentences. Our findings are consistent with those of \citet{keskar2016}, who also observe that the optimal batch size is at the lower end of the range, on a computer vision task with convolutional and fully-connected NNs.

\subsection{Asymptotic Performance}
\label{sec:asymptotic_performance}

An alternative explanation for the previous results is that the learning rate is too small for the larger batch sizes, which require more time to converge.  To test whether the differences observed above between small and large batch sizes depend on the actual training time, we continue training the 1,000 and 10,000 batch size models for HSB-DE and DE-EN for twice as many epochs as above (1400). The BLEU scores and their increases with respect to training for 700 epochs are given in Table~\ref{tab:bs_comparison_60k_400epochs}. The performance gap (from +3.85 to +3.25 BLEU) between small and large batch sizes is not overturned by training the models for much longer.

The scores from our best system (1,000 batch size, 42.81 BLEU on the test set) are similar to scores obtained by \emph{baselines} of the five highest-scoring teams at the WMT20 Low-resource shared task on HSB-DE \citep{fraser-2020-findings}.  While the scores of \citet{scherrer-etal-2020-university} and \citet{li-etal-2020-sjtu} are not comparable due to a different architecture or the use of unsupervised pre-training, the baseline scores of \citet{knowles-etal-2020-nrc-systems}, \citet{libovicky-etal-2020-lmu} and \citet{kvapilikova-etal-2020-cuni} are respectively 44.1, 43.4, and 38.7. The first one is higher than our best BLEU by 1.29, likely due to the use of 43M lines of CS and DE data for the subword vocabulary, vs.\ 700k in our case.

\begin{table}[ht]
\centering
\footnotesize
\begin{tabular}{lp{1cm}p{1cm}cp{1cm}p{1cm}}
\hline
    \textbf{Batch} & \multicolumn{2}{c}{\textbf{HSB-DE}} & & \multicolumn{2}{c}{\textbf{DE-EN}}\\ \cline{2-3} \cline{5-6}
    \textbf{size}                 & dev   & test & & dev   & test  \\ \hline
    1,000             & 49.52 (\textit{+0.29}) & 42.81 (\textit{+0.55}) & & 38.67 (\textit{+0.64}) & 39.24 (\textit{+0.57}) \\
    \hdashline
    10,000            & 46.44 (\textit{+1.83}) & 39.56 (\textit{+1.15}) & & 33.19 (\textit{+1.35}) & 34.42 (\textit{+0.92}) \\
\hline
\end{tabular}
\caption{BLEU scores for models trained for 1,400 epochs.  The scores for 1,000 are significantly higher (at 95\%) than those for 10,000.  In parenthesis, the absolute difference with BLEU scores after 700 epochs.}
\label{tab:bs_comparison_60k_400epochs}
\end{table}

\subsection{Invariance with respect to Vocabulary}

We additionally perform two comparisons that show that the above results hold regardless of the tokenizer and the vocabulary size. First, we test whether the score difference is preserved with an unshared SentencePiece vocabulary, i.e.\ when not sharing the source (HSB and DE) and the target (DE and EN) vocabularies. 

Second, we train two NMT models for HSB-DE using a Byte Pair Encoding (BPE) vocabulary \citep{P16-1162}, which we generate using the \texttt{learn\_bpe.py} tool from OpenNMT-py, with 32k merge operations and the remaining parameters at default values. Table~\ref{tab:bs_comparison_60k_vocabulary} shows BLEU scores on the development sets for batch sizes of 1,000 and 10,000. The previously observed differences in score between the batch sizes still hold, and we see that a shared Sentence\-Piece vocabulary leads to a better NMT system than an unshared or a BPE one.

\begin{table}[ht]
\centering
\footnotesize
\begin{tabular}{lp{1.2cm}p{1.2cm}p{1.2cm}}
\hline
    \textbf{Batch size} & \multicolumn{2}{c}{\textbf{SP unshared}} & \textbf{BPE}\\ 
    & \textbf{HSB-DE} & \textbf{DE-EN} & \textbf{HSB-DE} \\
    \hline
    1,000             & 46.80 (\textit{-2.43}) & 35.90 (\textit{-2.13}) & 46.21 (\textit{-3.02}) \\
    \hdashline
    10,000            & 41.99 (\textit{-2.62}) & 30.09 (\textit{-1.75}) & 43.35 (\textit{-1.26}) \\
\hline
\end{tabular}
\caption{BLEU scores on the dev set for HSB-DE and DE-EN models trained with SentencePiece (SP) vocabularies not shared between source and target (left) and BPE subwords (right).  The scores for 1,000 are significantly higher (at 95\%) than those for 10,000. In parenthesis, the difference with BLEU scores obtained with the SP shared vocabulary.}
\label{tab:bs_comparison_60k_vocabulary}
\end{table}


\section{Conclusion and Future Work}

In this work, we have shown that insights from computer vision on the regularizing effect of small batch sizes are also applicable to NMT.  Our results, focused on a low-resource setting, challenge those of previous NMT studies with large amounts of training data, and the general belief that batch sizes should be as large as they fit in the GPU memory. We have shown that training with small batch sizes leads to models that generalize better, and found the optimal batch size below which performance degrades. 

Future work should explore how the learning rate must be adjusted depending on the batch size, and whether a dynamically scheduled combination of batch size and learning rate can provide an even better regularizer. For instance, it should be tested if dynamic batch sizes as proposed by \citet{xu-etal-2020-dynamically} can also improve performance in a low-resource setting, with batch size thresholds changed to measure an optimal level of noise.


\section*{Acknowledgments}

We thank for their financial support the Swiss National Science Foundation (grant n.\ 175693 for the DOMAT project: On-demand Knowledge for Document-level Machine Translation) and Armasuisse (FamilyMT project).  We especially thank Dr.\ Ljiljana Dolamic (Armasuisse) for her support in the FamilyMT project.  We are also grateful to the anonymous reviewers and to Giorgos Vernikos for their helpful suggestions.

\bibliography{anthology,custom}

\begin{thebibliography}{30}
\expandafter\ifx\csname natexlab\endcsname\relax\def\natexlab#1{#1}\fi

\bibitem[{Araabi and Monz(2020)}]{araabi-monz-2020-optimizing}
Ali Araabi and Christof Monz. 2020.
\newblock \href {https://doi.org/10.18653/v1/2020.coling-main.304} {Optimizing
  {T}ransformer for low-resource neural machine translation}.
\newblock In \emph{Proceedings of the 28th International Conference on
  Computational Linguistics}, pages 3429--3435, Barcelona, Spain (Online).
  International Committee on Computational Linguistics.

\bibitem[{Bojar et~al.(2018)Bojar, Federmann, Fishel, Graham, Haddow, Koehn,
  and Monz}]{bojar-etal-2018-findings}
Ond{\v{r}}ej Bojar, Christian Federmann, Mark Fishel, Yvette Graham, Barry
  Haddow, Philipp Koehn, and Christof Monz. 2018.
\newblock \href {https://doi.org/10.18653/v1/W18-6401} {Findings of the 2018
  conference on machine translation ({WMT}18)}.
\newblock In \emph{Proceedings of the Third Conference on Machine Translation:
  Shared Task Papers}, pages 272--303, Belgium, Brussels. Association for
  Computational Linguistics.

\bibitem[{Dinh et~al.(2017)Dinh, Pascanu, Bengio, and Bengio}]{dinh2017sharp}
Laurent Dinh, Razvan Pascanu, Samy Bengio, and Yoshua Bengio. 2017.
\newblock Sharp minima can generalize for deep nets.
\newblock In \emph{International Conference on Machine Learning}, pages
  1019--1028. PMLR.

\bibitem[{Fraser(2020)}]{fraser-2020-findings}
Alexander Fraser. 2020.
\newblock \href {https://aclanthology.org/2020.wmt-1.80} {Findings of the {WMT}
  2020 shared tasks in unsupervised {MT} and very low resource supervised
  {MT}}.
\newblock In \emph{Proceedings of the Fifth Conference on Machine Translation},
  pages 765--771, Online. Association for Computational Linguistics.

\bibitem[{Goodfellow et~al.(2016)Goodfellow, Bengio, and
  Courville}]{Goodfellow-et-al-2016}
Ian Goodfellow, Yoshua Bengio, and Aaron Courville. 2016.
\newblock \emph{Deep Learning}.
\newblock MIT Press.

\bibitem[{Goyal et~al.(2017)Goyal, Doll{\'a}r, Girshick, Noordhuis, Wesolowski,
  Kyrola, Tulloch, Jia, and He}]{goyal2018accurate}
Priya Goyal, Piotr Doll{\'a}r, Ross Girshick, Pieter Noordhuis, Lukasz
  Wesolowski, Aapo Kyrola, Andrew Tulloch, Yangqing Jia, and Kaiming He. 2017.
\newblock Accurate, large minibatch {SGD}: Training {I}magenet in 1 hour.
\newblock \emph{arXiv preprint arXiv:1706.02677}.

\bibitem[{Hoffer et~al.(2017)Hoffer, Hubara, and
  Soudry}]{hoffer_trainlonger2017}
Elad Hoffer, Itay Hubara, and Daniel Soudry. 2017.
\newblock Train longer, generalize better: Closing the generalization gap in
  large batch training of neural networks.
\newblock In \emph{Proceedings of the 31st International Conference on Neural
  Information Processing Systems}, NIPS'17, page 1729–1739, Red Hook, NY,
  USA. Curran Associates Inc.

\bibitem[{Kawaguchi et~al.(2017)Kawaguchi, Kaelbling, and
  Bengio}]{kawaguchi2020generalization}
Kenji Kawaguchi, Leslie~Pack Kaelbling, and Yoshua Bengio. 2017.
\newblock Generalization in deep learning.
\newblock \emph{arXiv preprint arXiv:1710.05468}.

\bibitem[{Keskar et~al.(2016)Keskar, Mudigere, Nocedal, Smelyanskiy, and
  Tang}]{keskar2016}
Nitish~Shirish Keskar, Dheevatsa Mudigere, Jorge Nocedal, Mikhail Smelyanskiy,
  and Ping Tak~Peter Tang. 2016.
\newblock On large-batch training for deep learning: Generalization gap and
  sharp minima.
\newblock \emph{arXiv preprint arXiv:1609.04836}.

\bibitem[{Klein et~al.(2020)Klein, Hernandez, Nguyen, and
  Senellart}]{klein-etal-2020-opennmt}
Guillaume Klein, Fran{\c{c}}ois Hernandez, Vincent Nguyen, and Jean Senellart.
  2020.
\newblock \href {https://aclanthology.org/2020.amta-research.9} {The
  {O}pen{NMT} neural machine translation toolkit: 2020 edition}.
\newblock In \emph{Proceedings of the 14th Conference of the Association for
  Machine Translation in the Americas (Volume 1: Research Track)}, pages
  102--109, Virtual. Association for Machine Translation in the Americas.

\bibitem[{Klein et~al.(2017)Klein, Kim, Deng, Senellart, and Rush}]{P17-4012}
Guillaume Klein, Yoon Kim, Yuntian Deng, Jean Senellart, and Alexander Rush.
  2017.
\newblock \href {http://aclweb.org/anthology/P17-4012} {Open{NMT}: Open-source
  toolkit for {NMT}}.
\newblock In \emph{Proceedings of ACL 2017, System Demonstrations}, pages
  67--72. Association for Computational Linguistics.

\bibitem[{Knowles et~al.(2020)Knowles, Larkin, Stewart, and
  Littell}]{knowles-etal-2020-nrc-systems}
Rebecca Knowles, Samuel Larkin, Darlene Stewart, and Patrick Littell. 2020.
\newblock \href {https://aclanthology.org/2020.wmt-1.132} {{NRC} systems for
  low resource {G}erman-{U}pper {S}orbian machine translation 2020: Transfer
  learning with lexical modifications}.
\newblock In \emph{Proceedings of the Fifth Conference on Machine Translation},
  pages 1112--1122, Online. Association for Computational Linguistics.

\bibitem[{Kudo(2018)}]{Kudo2018-xx}
Taku Kudo. 2018.
\newblock Subword regularization: Improving neural network translation models
  with multiple subword candidates.
\newblock In \emph{Proceedings of the 56th Annual Meeting of the Association
  for Computational Linguistics (Volume 1: Long Papers)}, pages 66--75.

\bibitem[{Kvapil{\'\i}kov{\'a} et~al.(2020)Kvapil{\'\i}kov{\'a}, Kocmi, and
  Bojar}]{kvapilikova-etal-2020-cuni}
Ivana Kvapil{\'\i}kov{\'a}, Tom Kocmi, and Ond{\v{r}}ej Bojar. 2020.
\newblock \href {https://aclanthology.org/2020.wmt-1.133} {{CUNI} systems for
  the unsupervised and very low resource translation task in {WMT}20}.
\newblock In \emph{Proceedings of the Fifth Conference on Machine Translation},
  pages 1123--1128, Online. Association for Computational Linguistics.

\bibitem[{Li et~al.(2017)Li, Xu, Taylor, Studer, and
  Goldstein}]{li2018visualizing}
Hao Li, Zheng Xu, Gavin Taylor, Christoph Studer, and Tom Goldstein. 2017.
\newblock Visualizing the loss landscape of neural nets.
\newblock \emph{arXiv preprint arXiv:1712.09913}.

\bibitem[{Li et~al.(2020)Li, Zhao, Wang, Chen, Utiyama, and
  Sumita}]{li-etal-2020-sjtu}
Zuchao Li, Hai Zhao, Rui Wang, Kehai Chen, Masao Utiyama, and Eiichiro Sumita.
  2020.
\newblock \href {https://aclanthology.org/2020.wmt-1.22} {{SJTU}-{NICT}{'}s
  supervised and unsupervised neural machine translation systems for the
  {WMT}20 news translation task}.
\newblock In \emph{Proceedings of the Fifth Conference on Machine Translation},
  pages 218--229, Online. Association for Computational Linguistics.

\bibitem[{Libovick{\'y} et~al.(2020)Libovick{\'y}, Hangya, Schmid, and
  Fraser}]{libovicky-etal-2020-lmu}
Jind{\v{r}}ich Libovick{\'y}, Viktor Hangya, Helmut Schmid, and Alexander
  Fraser. 2020.
\newblock \href {https://aclanthology.org/2020.wmt-1.131} {The {LMU} {M}unich
  system for the {WMT}20 very low resource supervised {MT} task}.
\newblock In \emph{Proceedings of the Fifth Conference on Machine Translation},
  pages 1104--1111, Online. Association for Computational Linguistics.

\bibitem[{Libovický and Fraser(2021)}]{fraser-2021-findings}
Jindřich Libovický and Alexander Fraser. 2021.
\newblock \href {http://www.statmt.org/wmt21/pdf/2021.wmt-1.72.pdf} {Findings
  of the {WMT} 2021 shared tasks in unsupervised {MT} and very low resource
  supervised {MT}}.
\newblock In \emph{Proceedings of the Sixth Conference on Machine Translation},
  Online. Association for Computational Linguistics.

\bibitem[{Papineni et~al.(2002)Papineni, Roukos, Ward, and
  Zhu}]{papineni-etal-2002-bleu}
Kishore Papineni, Salim Roukos, Todd Ward, and Wei-Jing Zhu. 2002.
\newblock \href {https://doi.org/10.3115/1073083.1073135} {{B}leu: a method for
  automatic evaluation of machine translation}.
\newblock In \emph{Proceedings of the 40th Annual Meeting of the Association
  for Computational Linguistics}, pages 311--318, Philadelphia, Pennsylvania,
  USA. Association for Computational Linguistics.

\bibitem[{Popel and Bojar(2018)}]{Popel_2018_tips}
Martin Popel and Ondřej Bojar. 2018.
\newblock \href {https://doi.org/10.2478/pralin-2018-0002} {Training tips for
  the {T}ransformer model}.
\newblock \emph{The Prague Bulletin of Mathematical Linguistics},
  110(1):43–70.

\bibitem[{Popovi{\'c}(2015)}]{popovic-2015-chrf}
Maja Popovi{\'c}. 2015.
\newblock \href {https://doi.org/10.18653/v1/W15-3049} {chr{F}: character
  n-gram f-score for automatic {MT} evaluation}.
\newblock In \emph{Proceedings of the Tenth Workshop on Statistical Machine
  Translation}, pages 392--395, Lisbon, Portugal. Association for Computational
  Linguistics.

\bibitem[{Post(2018)}]{post-2018-call}
Matt Post. 2018.
\newblock \href {https://doi.org/10.18653/v1/W18-6319} {A call for clarity in
  reporting {BLEU} scores}.
\newblock In \emph{Proceedings of the Third Conference on Machine Translation:
  Research Papers}, pages 186--191, Belgium, Brussels. Association for
  Computational Linguistics.

\bibitem[{Scherrer et~al.(2020)Scherrer, Gr{\"o}nroos, and
  Virpioja}]{scherrer-etal-2020-university}
Yves Scherrer, Stig-Arne Gr{\"o}nroos, and Sami Virpioja. 2020.
\newblock \href {https://aclanthology.org/2020.wmt-1.134} {The {U}niversity of
  {H}elsinki and {A}alto university submissions to the {WMT} 2020 news and
  low-resource translation tasks}.
\newblock In \emph{Proceedings of the Fifth Conference on Machine Translation},
  pages 1129--1138, Online. Association for Computational Linguistics.

\bibitem[{Sennrich et~al.(2016)Sennrich, Haddow, and Birch}]{P16-1162}
Rico Sennrich, Barry Haddow, and Alexandra Birch. 2016.
\newblock \href {https://doi.org/10.18653/v1/P16-1162} {{NMT} of rare words
  with subword units}.
\newblock In \emph{Proceedings of the 54th Annual Meeting of the Association
  for Computational Linguistics (Volume 1: Long Papers)}, pages 1715--1725.

\bibitem[{Sennrich and Zhang(2019)}]{sennrich-zhang-2019-revisiting}
Rico Sennrich and Biao Zhang. 2019.
\newblock \href {https://doi.org/10.18653/v1/P19-1021} {Revisiting low-resource
  neural machine translation: A case study}.
\newblock In \emph{Proceedings of the 57th Annual Meeting of the Association
  for Computational Linguistics}, pages 211--221, Florence, Italy. Association
  for Computational Linguistics.

\bibitem[{Snover et~al.(2006)Snover, Dorr, Schwartz, Micciulla, and
  Makhoul}]{snover-2006-ter}
Matthew Snover, Bonnie Dorr, Richard Schwartz, Linnea Micciulla, and John
  Makhoul. 2006.
\newblock \href {http://mt-archive.info/AMTA-2006-Snover.pdf} {A study of
  translation edit rate with targeted human annotations}.
\newblock In \emph{Proceedings of the Conference of the Association for Machine
  Translation in the Americas (AMTA 2006)}.

\bibitem[{Tiedemann(2012)}]{TIEDEMANN12.463}
Jörg Tiedemann. 2012.
\newblock Parallel data, tools and interfaces in {OPUS}.
\newblock In \emph{Proceedings of the Eight International Conference on
  Language Resources and Evaluation (LREC'12)}, Istanbul, Turkey. European
  Language Resources Association (ELRA).

\bibitem[{Van~Biljon et~al.(2020)Van~Biljon, Pretorius, and
  Kreutzer}]{van2020optimal}
Elan Van~Biljon, Arnu Pretorius, and Julia Kreutzer. 2020.
\newblock On optimal {T}ransformer depth for low-resource language translation.
\newblock \emph{arXiv preprint arXiv:2004.04418}.

\bibitem[{Vaswani et~al.(2017)Vaswani, Shazeer, Parmar, Uszkoreit, Jones,
  Gomez, Kaiser, and Polosukhin}]{vaswani2017_attention}
Ashish Vaswani, Noam Shazeer, Niki Parmar, Jakob Uszkoreit, Llion Jones,
  Aidan~N Gomez, Lukasz Kaiser, and Illia Polosukhin. 2017.
\newblock \href
  {http://papers.nips.cc/paper/7181-attention-is-all-you-need.pdf} {Attention
  is all you need}.
\newblock In \emph{Advances in Neural Information Processing Systems 30}, pages
  5998--6008.

\bibitem[{Xu et~al.(2020)Xu, van Genabith, Xiong, and
  Liu}]{xu-etal-2020-dynamically}
Hongfei Xu, Josef van Genabith, Deyi Xiong, and Qiuhui Liu. 2020.
\newblock \href {https://doi.org/10.18653/v1/2020.acl-main.323} {Dynamically
  adjusting {T}ransformer batch size by monitoring gradient direction change}.
\newblock In \emph{Proceedings of the 58th Annual Meeting of the Association
  for Computational Linguistics}, pages 3519--3524, Online. Association for
  Computational Linguistics.

\end{thebibliography}
\bibliographystyle{acl_natbib}

\appendix

\section{Appendix}
\label{sec:appendix}

The hyper-parameters used to train our models are the following ones:

\noindent \texttt{src\_words\_min\_frequency: 2}\\
\texttt{tgt\_words\_min\_frequency: 2}\\
\texttt{valid\_batch\_size: 200}\\
\texttt{max\_generator\_batches: 2}\\
\texttt{optim: adam}\\
\texttt{learning\_rate: 2.0}\\
\texttt{adam\_beta2: 0.998}\\
\texttt{decay\_method: noam}\\
\texttt{accum\_count: 2}\\
\texttt{warmup\_steps: 8000}\\
\texttt{label\_smoothing: 0.1}\\
\texttt{max\_grad\_norm: 0}\\
\texttt{param\_init: 0}\\
\texttt{param\_init\_glorot: true}\\
\texttt{normalization: tokens}\\
\texttt{encoder\_type: transformer}\\
\texttt{decoder\_type: transformer}\\
\texttt{position\_encoding: true}\\
\texttt{layers: 6}\\
\texttt{heads: 8} \\
\texttt{rnn\_size: 512}\\
\texttt{word\_vec\_size: 512}\\
\texttt{transformer\_ff: 2048}\\
\texttt{dropout: 0.1}\\
\texttt{batch\_size: 1000}\\
\texttt{batch\_type: tokens}\\

\end{document}